\documentclass[journal]{IEEEtran}

\usepackage{epsfig, amsmath, amssymb, cite, booktabs, array, multirow}
\usepackage{dblfloatfix} 
\usepackage{arydshln}
\usepackage{mathtools}
\usepackage{mathrsfs}
\usepackage{boldline}
\usepackage{algorithm}
\usepackage{algorithmic}
\usepackage{multirow}
\usepackage[normalem]{ulem}
\usepackage{graphicx}
\usepackage{enumitem}
\usepackage{graphicx}
\usepackage{makecell}
\usepackage{color}
\usepackage{subfigure}
\useunder{\uline}{\ul}{}
\makeatletter
\newcommand*{\rom}[1]{\expandafter\@slowromancap\romannumeral #1@}
\makeatother

\graphicspath{{./figure/}}

\begin{document}

\title{Oracle Teacher: Leveraging Target Information for Better Knowledge Distillation of CTC Models}
\author{Ji~Won~Yoon, 
Hyung~Yong~Kim,
Hyeonseung~Lee,
Sunghwan~Ahn, and Nam~Soo~Kim

\thanks{Manuscript received 22 October 2022; revised 13 March 2023; accepted 5
July 2023. Date of publication 21 July 2023; date of current version 10 August
2023. The associate editor coordinating the review of this
manuscript and approving it for publication was Dr. Yu Zhang. \emph{(Corresponding
author: Nam Soo Kim.)}}
\thanks{Ji Won Yoon, Hyeonseung Lee, Sunghwan Ahn, and Nam Soo Kim are with
the Department of Electrical and Computer Engineering and the Institute of New
Media and Communications, Seoul National University, Seoul 08826, South
Korea (e-mail: jwyoon@hi.snu.ac.kr; hslee@hi.snu.ac.kr; shahn@hi.snu.ac.kr;
nkim@snu.ac.kr).}
\thanks{Hyung Yong Kim was with the Seoul National University, Seoul 08826,
South Korea. He is now with the 42dot Inc., Seoul 06620, South Korea (e-mail:
hyungyong.kim@42dot.ai).}
\thanks{Digital Object Identifier 10.1109/TASLP.2023.3297955}
}


\ifCLASSOPTIONpeerreview
\author{\IEEEauthorblockN{Ji~Won~Yoon,~\IEEEmembership{Student Member,~IEEE} and Nam~Soo~Kim,~\IEEEmembership{Senior Member,~IEEE}}
\IEEEauthorblockA{Department of Electrical and Computer Engineering and INMC,\\
Seoul National University \\
1 Gwanak-ro, Gwanak-gu, Seoul 08826, Korea \\
Tel: +82-2-880-8439 Fax: +82-2-880-8219 E-mail:
nkim@snu.ac.kr}} \fi


\maketitle

\begin{abstract}
Knowledge distillation (KD), best known as an effective method for model compression, aims at transferring the knowledge of a bigger network (teacher) to a much smaller network (student). Conventional KD methods usually employ the teacher model trained in a supervised manner, where output labels are treated only as targets. Extending this supervised scheme further, we introduce a new type of teacher model for connectionist temporal classification (CTC)-based sequence models, namely Oracle Teacher, that leverages both the source inputs and the output labels as the teacher model's input. Since the Oracle Teacher learns a more accurate CTC alignment by referring to the target information, it can provide the student with more optimal guidance. One potential risk for the proposed approach is a trivial solution that the model's output directly copies the target input. Based on a many-to-one mapping property of the CTC algorithm, we present a training strategy that can effectively prevent the trivial solution and thus enables utilizing both source and target inputs for model training. Extensive experiments are conducted on two sequence learning tasks: speech recognition and scene text recognition. From the experimental results, we empirically show that the proposed model improves the students across these tasks while achieving a considerable speed-up in the teacher model's training time.
\end{abstract}

\begin{IEEEkeywords}
Speech recognition, scene text recognition, connectionist temporal classification, knowledge distillation, teacher-student learning, transfer learning
\end{IEEEkeywords}

\IEEEpeerreviewmaketitle

\section{Introduction}

\IEEEPARstart{A}s deep neural networks bring a significant improvement in various fields such as speech recognition, computer vision, and natural language processing, they also become wider and deeper.
However, as models grow in size and complexity, high-performing neural network models become either computationally expensive or consume a large amount of memory, hindering their wide deployment in resource-limited scenarios.
To mitigate this computational burden, several techniques such as model pruning \cite{pruning1:scheme, pruning2:scheme}, quantization\cite{quantization:scheme}, and knowledge distillation \cite{bucila-et-al:scheme, hinton_kd-et-al:scheme} have been suggested.
Among these approaches, knowledge distillation (KD) is a popular compression scheme, which is the process of transferring knowledge from a deep and complex model (teacher) to a shallower and simpler model (student). 

Conventional KD methods \cite{hinton_kd-et-al:scheme, romero-et-al:scheme, kurata-et-al:scheme, kurata2-et-al:scheme, takashima-et-al:scheme, takashima-et-al2:scheme, tutornet:scheme} typically share a common feature; they require a teacher model with high capacity that has been trained in a supervised manner, where the ground-truth labels are required as a target.
However, training the teacher from scratch can be costly since many of the current state-of-the-art (SOTA) models suffer from
excessive training time and difficult hyper-parameters tuning.
Thus, some existing approaches \cite{sec1_add1,sec1_add2,sec1_add3} rely heavily on the pre-trained model, provided by other prior research, as the teacher to save the training time and resource cost.
Even though making full use of the provided pre-trained models is one important motivation of KD, this dependency might limit the flexibility of our consideration.
If we can train a better teacher model with fewer resources and training time, KD from various teachers will be possible on different tasks or databases.

We revisit the teacher model in KD from a different perspective.
In a conventional KD scenario, there is no guarantee that the teacher can find the correct solution for every complicated problem in an optimal way, implying that the teacher model may provide suboptimal guidance for the student.
The key idea of our framework is to derive a more accurate problem-solving process by referring to the existing solutions so that the teacher can provide better guidance to the student.
On this basis, we introduce a new type of teacher for Connectionist Temporal Classification (CTC) \cite{graves-et-al:scheme}-based sequence models, namely Oracle Teacher.
The conventional teacher is typically built in a supervised manner whose goal is to predict the target output for a given source input data. 
In contrast, the proposed teacher model utilizes not only the source input but also the target value to estimate better CTC alignment.

However, it may be somewhat confusing to understand what it means to train a model using both the source inputs and the output labels as the model's input.
Specifically, the Oracle Teacher is likely to heavily rely on the target input, i.e., the output label, while ignoring the embedding from the source input.
To overcome this problem, we propose a training scheme that uses the many-to-one mapping property of the CTC algorithm.
Since the relationship between the CTC alignment and the original target is many-to-one, 
we can prevent a trivial solution that the model's output directly copies the target input.
To the best of our knowledge, this is the first attempt of using the target input to improve the ability of the teacher model.
Utilizing the target input for training the teacher model brings several benefits for KD. Firstly, the proposed teacher model produces a more accurate CTC alignment by referring to the target information so that its knowledge can provide more optimal guidance to the student.
Secondly, the representation of the proposed teacher contains target-related embedding that can be supportive for student training. 
For example, the Oracle Teacher for automatic speech recognition (ASR) is trained to use both speech and text as the model's input during training.
Different from the typical ASR teachers that take only acoustic features into consideration, the Oracle Teacher performs a fusion of both acoustic (speech) and linguistic (text) features when generating the prediction.
Since unifying acoustic and linguistic representation learning generally enhances the performance of the speech processing \cite{acours_lin1,acours_lin2,acours_lin3,acours_lin4,acours_lin5}, the Oracle Teacher's representation, which considers not only the acoustic but also linguistic information, can be more effective for the ASR student.
Also, the Oracle Teacher can boost up the speed of the training since the target input is used as the guidance to reduce the candidate scope of the prediction. Compared to the conventional teacher models that require tremendous time and GPU resources, our framework dramatically reduces the computational cost required to train the teacher model.

Extensive experiments are conducted on two different sequence learning tasks: ASR and scene text recognition (STR). 
Empirically, we verify that the student distilled from the Oracle Teacher achieves better performance compared to the case when it is distilled from the other pre-trained models, which yield the high performance for each task.
Apart from performance, we measure the computational cost for training teacher model and show that a powerful teacher can be trained with a reduced computational burden via the proposed scheme.
Through an in-depth case study, we also analyze the effect of target injection and the linguistic information acquired from the encoder.
Additionally, it is verified that Oracle Teacher performs well in different KD scenarios, including a transducer framework and a self-training setting.

Our \textbf{main contributions} are summarized as follows:
\begin{enumerate}
    \item  Our paper introduces a new type of teacher for CTC-based sequence models, namely Oracle Teacher, that utilizes the output labels as an additional input for model training.
    The proposed teacher model can estimate a more accurate CTC alignment, providing more optimal guidance to the student.
    To the best of our knowledge, this is the first attempt of using the target input to improve the performance of a teacher model.
    \item Through extensive experiments on two sequence learning tasks, including ASR and STR, we verify the superiority of the Oracle Teacher compared to the conventional teacher models.
    Moreover, our framework dramatically reduces the computational cost of the teacher model in terms of the training time and required GPU resources.
    \item In a detailed case study and analysis, we validate why the proposed method can result in better KD performance than the conventional teacher and check if the Oracle Teacher is correctly trained while preventing the trivial solution.
    
\end{enumerate}

\section{Related work}
\label{2}
\subsection{Knowledge Distillation}
\label{2.1}
There has been a long line of research on KD, which aims at distilling knowledge from a big teacher model to a small student model. 
Bucila \textit{et al.} \cite{bucila-et-al:scheme} proposed a method to compress an ensemble of models into a single model without significant accuracy loss.
Later, Ba and Caruana \cite{dodeep:scheme} extended it to deep learning by using the logits of the teacher model.
Hinton \textit{et al.} \cite{hinton_kd-et-al:scheme} revived this idea under the name of KD that distills class probability by minimizing the Kullback-Leibler (KL)-divergence between the softmax outputs of the teacher and student.
In the case of the ASR task, the most frequently employed KD approach is to train a student with the teacher's prediction as a target, in conjunction with the ground truth.
For the conventional deep neural network (DNN)-hidden Markov model (HMM) hybrid systems, Li \textit{et al.} \cite{firstasr:scheme} first attempted to apply the teacher-student learning to a speech recognition task, and Wong \textit{et al.} \cite{seq:scheme} applied sequence-level KD to the acoustic model. Several researchers applied KD to improve the performance by minimizing the frame-level cross-entropy loss between the output distributions of the teacher and student \cite{blending:scheme,chebotar-et-al:scheme,watanabe-et-al:scheme,lu-et-al:scheme,fukuda-et-al:scheme}.
For end-to-end speech recognition, KD has been successfully applied to CTC models \cite{senior-et-al:scheme,takashima-et-al:scheme, takashima-et-al2:scheme, kurata2-et-al:scheme,kurata-et-al:scheme, tutornet:scheme} and attention-based encoder-decoder models \cite{compression:scheme, wer_error:scheme, entropy:scheme, tutornet:scheme}.
However, as reported in previous KD studies \cite{senior-et-al:scheme,takashima-et-al:scheme, takashima-et-al2:scheme, tutornet:scheme}, simply applying the frame-level CE to the CTC-based model can worsen the performance compared to the baseline.
To cover this problem, Kurata and Audhkhasi \cite{kurata-et-al:scheme, kurata2-et-al:scheme} proposed KD approaches, where the CTC-based student can be trained using the frame-wise alignment of the teacher.
Takashima \textit{et al.} \cite{takashima-et-al:scheme,takashima-et-al2:scheme} explored sequence-level KD methods for training CTC models.
Yoon \textit{et al.} \cite{tutornet:scheme} suggested that $l_{2}$ loss is more suitable than the conventional KL-divergence to distill frame-level posterior in the CTC framework.
Moritz \textit{et al.} \cite{gtc} newly proposed graph-based temporal classification (GTC) objective, which is applied for self-training with WFST-based supervision.

The hidden representation from the teacher also has been proven to hold additional knowledge that can contribute to improving the student's performance.
Recently, some KD methods \cite{romero-et-al:scheme,at:scheme,fsp:scheme,jacobian:scheme,factor:scheme, boundary:scheme, relation:scheme, showanddistill:scheme}, particularly in computer vision, were proposed to minimize the mean squared error (MSE) between the representation-level knowledge of the two models.
They address how to extract a better knowledge from the teacher model and transfer it to the student.
Yoon \textit{et al.} \cite{tutornet:scheme} first attempted to transfer the the hidden representation across different structured neural networks for end-to-end speech recognition while using frame weighting that reflects which frames are important for KD.
Recently, several KD approaches \cite{distillhubert:scheme, lighthubert:scheme,fithubert} suggested using the hidden representation-level knowledge to improve the self-supervised speech representation learning-based models, like Hidden-Unit BERT (HuBERT).

\subsection{Connectionist Temporal Classification}
\label{2.2}
Generally, an end-to-end sequence model directly converts a sequence of input features $x_{1:T}$ into a sequence of target labels $y_{1:L}$ where $y_{l} \in \mathcal{I}$ with $\mathcal{I}$ being the set of labels. $T$ and $L$ are respectively the length of $x=x_{1:T}$ and $y=y_{1:L}$. 
To cope with the mapping problem when the two sequences have different lengths, the Connectionist Temporal Classification (CTC) framework \cite{graves-et-al:scheme} introduces~``blank" label and allows the repetition of each label to force the output and input sequences to have the same length. A CTC alignment $\pi_{1:T}$ is a sequence of initial output labels, as every input $x_{t}$ is mapped to a certain label $\pi_{t} \in \mathcal{I'}$ where $\mathcal{I'} = \mathcal{I} \cup \{blank\}$. A mapping function $\mathcal{B}$, which is defined as $y = \mathcal{B}(\pi)$, maps the alignment sequence $\pi$ into the final output sequence $y$ after merging consecutive repeated characters and removing blank labels. 
The conditional probability of the target sequence $y$ given the input sequence $x$ is defined as
\begin{equation}
   P(y \vert x) = \sum_{\pi \in \mathcal{B}^{-1}(y)} P(\pi \vert x).
   \label{ctc}
\end{equation}
where $\mathcal{B}^{-1}$ denotes the inverse mapping and returns all possible alignment sequences compatible with $y$.
Given a target label sequence $y$, the loss function $\mathcal{L}_{CTC}$ is defined as:
\begin{align}
  \mathcal{L}_{CTC} = - \log P(y \vert x) .
\end{align}%

\subsection{Recurrent Neural Network Transducer}
\label{2.3}
An alternative approach to the end-to-end mapping between $x_{1:T}$ and $y_{1:L}$ is to use the recurrent neural network transducer (RNN-T) \cite{graves-rnn-t:scheme}. 
The RNN-T model typically consists of the encoder network, the joint network, and the prediction network. The encoder network generates the acoustic embedding from the speech source, and the prediction network processes labels independent of the acoustics.
The joint network integrates the outputs of the encoder and prediction networks in generating the prediction.

\subsection{Text Injection ASR Model}
\label{2.4}
There are some techniques in ASR literature that use both speech and text inputs.
Chen \textit{et al.} \cite{maestro} proposed Maestro, which is a self-supervised training method that learns unified representations from both speech and text. 
Thomas \textit{et al.} \cite{textogram} suggested Textogram that integrates the text input for training the ASR model.
It employs a concatenation of the one-hot encodings of the symbols making up the reference text.
In contrast to the conventional KD framework that improves the teacher model through back-propagation of a loss based on the target, the proposed Oracle Teacher leverages the target through forward propagation to improve the teacher model, similar to other text-injection approaches.

\section{Oracle Teacher}
\begin{figure}[t]
\centering
	\includegraphics[height=10cm]{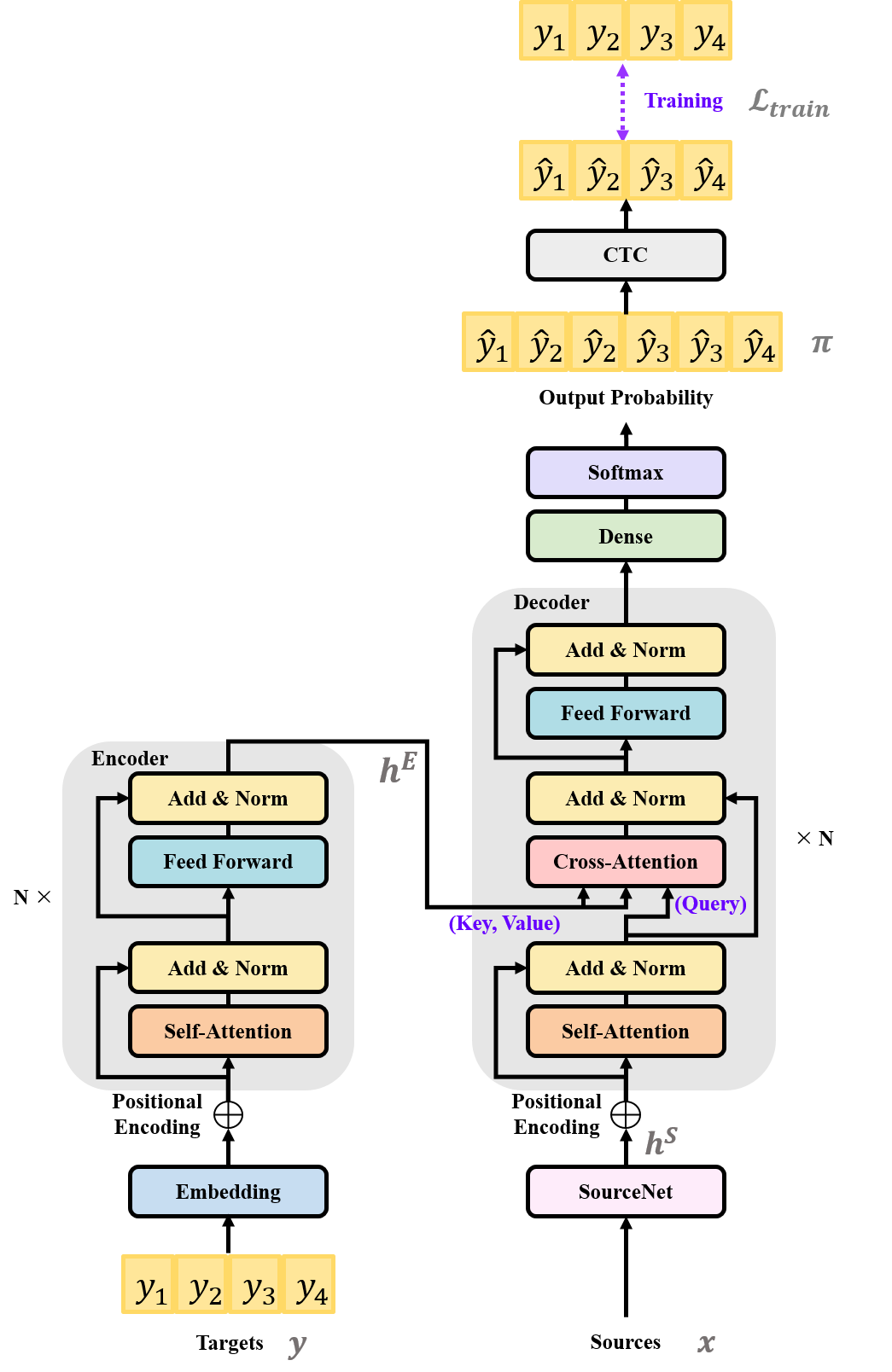}
	\caption{Overview of the Oracle Teacher. The proposed teacher model mainly consists of three components: the SourceNet, the encoder, and the decoder. Different from the conventional teacher, the target $y$ is used as the additional input to the model. Note that the Oracle Teacher is a non-autoregressive model where the look-ahead mask is not included in the decoder. The architecture selection of the SourceNet depends on the task we are interested in. When the main task is ASR. the SourceNet corresponds to an acoustic model part of the conventional ASR model. In our experiment for ASR, the SourceNet is based on the architecture of Japser \cite{jasper:scheme}. For STR, we apply the CRNN \cite{crnn:scheme} as the SourceNet.}
	\label{oracle_arc}
\end{figure}

This section introduces how to design the Oracle teacher that utilizes the output labels as an additional input.
As shown in Fig. \ref{oracle_arc}, we let the Oracle Teacher model learn a function from the source $x$ and the target $y$ inputs to the CTC alignment $\pi$.

\begin{figure}[t]
    \centering
    \includegraphics[height=3.1cm]{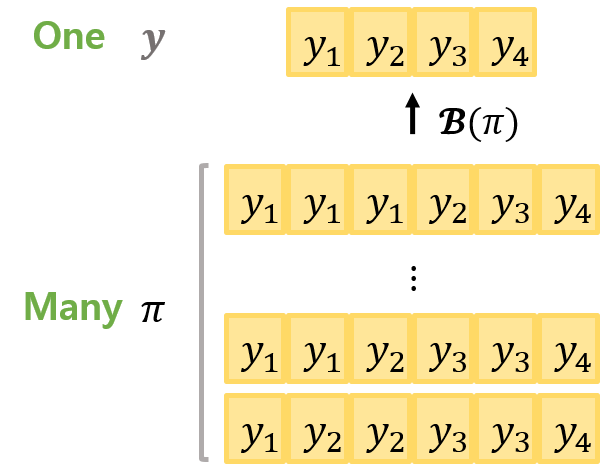}
    \caption{The relationship between the CTC alignment $\pi$ and the target input $y$. A many-to-one mapping function $\mathcal{B}$ converts the alignment sequence $\pi$ into the final output sequence $y$.}
    \label{CTC_MANY}
\end{figure}

\subsection{Oracle Teacher Training}
Let $x=x_{1:T}=\{x_{1},...,x_{T}\}$ be an input sequence of length $T$, and $y=y_{1:L}=\{y_{1},...,y_{L}\}$ be a target sequence of length $L$.
As mentioned in Section \ref{2.2},  the CTC algorithm employs the intermediate CTC alignment $\pi=\pi_{1:T}=\{\pi_{1},...,\pi_{T}\}$ to align variable-length input and output sequences.
Note that the initial output of the CTC model is $\pi$.
As shown in Fig. \ref{CTC_MANY}, the relationship between the alignment $\pi$ and the target input $y$ is many-to-one via the mapping $\mathcal{B}$.
Let us consider the single word ``cat", there are many possible alignment sequences compatible with y, e.g., $\mathcal{B}(\{blank, c, blank, a, t, t\})= \mathcal{B}(\{c, c, blank, a, blank, t\})= \mathcal{B}(\{blank, blank, blank, c, a, t\})=\{cat\}$ where $\{cat\}$ denotes $y$, and the other sequences represent $\pi$.
This many-to-one setting is the key to training the Oracle Teacher while preventing the trivial solution.
Intuitively, predicting the CTC alignment $\pi$ (many) from the target input $y$ (one) should be difficult since many possible paths are compatible with $y$.
To generate an accurate CTC prediction, the linguistic information from $\{cat\}$ is not enough.
The model needs to be trained to assign suitable text information for each frame, like \{c, c, blank, a, blank, t\}. 
Since aligning linguistic information to each frame is related to the acoustic information, the proposed model is trained to use the embeddings of both $x$ and $y$.
The detailed analysis will be described in Section \ref{linguistic}.

The Oracle Teacher learns the parameters $\theta$ to minimize the following training loss:
\begin{equation}
\label{oracle}
\mathcal{L}_{train} = -\log\sum_{\pi \in \mathcal{B}^{-1}(y)}P(\pi|x,y;\theta)
\end{equation}
where $\mathcal{B}$ is the many-to-one mapping function in (\ref{ctc}) that maps the latent alignment $\pi_{1:T}$ into the target $y_{1:L}$.

\subsection{Knowledge Distillation with Oracle Teacher} 
\label{oracle_kd_section}
We can interpret KD framework from a different perspective by applying the additional target input.
Given a source $x$, the student model learns the parameter $\phi$ to maximize the following conditional probability:
\begin{align}
\label{kl}
  \log P(y|x;\phi) & = \log\sum_{\pi}P(y,\pi|x;\phi)\,d\pi  \nonumber \\
    & = \log\sum_{\pi}P(y,\pi|x;\phi)\frac{P(\pi|x,y;\theta)}{P(\pi|x,y;\theta)}\,d\pi \nonumber \\
  & = \log\sum_{\pi}\delta(y-\mathcal{B}(\pi))P(\pi|x;\phi)\frac{P(\pi|x,y;\theta)}{P(\pi|x,y;\theta)}\,d\pi \nonumber \\
 & = \log\sum_{\pi \in \mathcal{B}^{-1}(y)}P(\pi|x,y;\theta)\frac{P(\pi|x;\phi)}{P(\pi|x,y;\theta)}\,d\pi \nonumber \\
  & \geq \sum_{\pi \in \mathcal{B}^{-1}(y)}P(\pi|x,y;\theta)\log\frac{P(\pi|x;\phi)}{P(\pi|x,y;\theta)}\,d\pi  \nonumber \\
  & = - D_{KL}(\underbrace{P(\pi|x,y;\theta)}_{\text{Oracle Teacher}} \Vert \underbrace{P(\pi|x;\phi)}_{\text{Student}}) 
\end{align}%
where the inequality follows from Jensen’s inequality, $\mathcal{B}$ represents the mapping function in (\ref{oracle}), and $D_{KL}$ denotes the KL-divergence. 
In our framework, $P(\pi|x,y;\theta)$ and $P(\pi|x;\phi)$ correspond to the alignment probability derived from the Oracle Teacher and the student, respectively.
By minimizing the KL-divergence between the Oracle Teacher and the student, we can maximize the conditional probability of the student model $P(y|x;\phi)$.

\begin{figure}[t]
\centering
	\includegraphics[height=4.7cm]{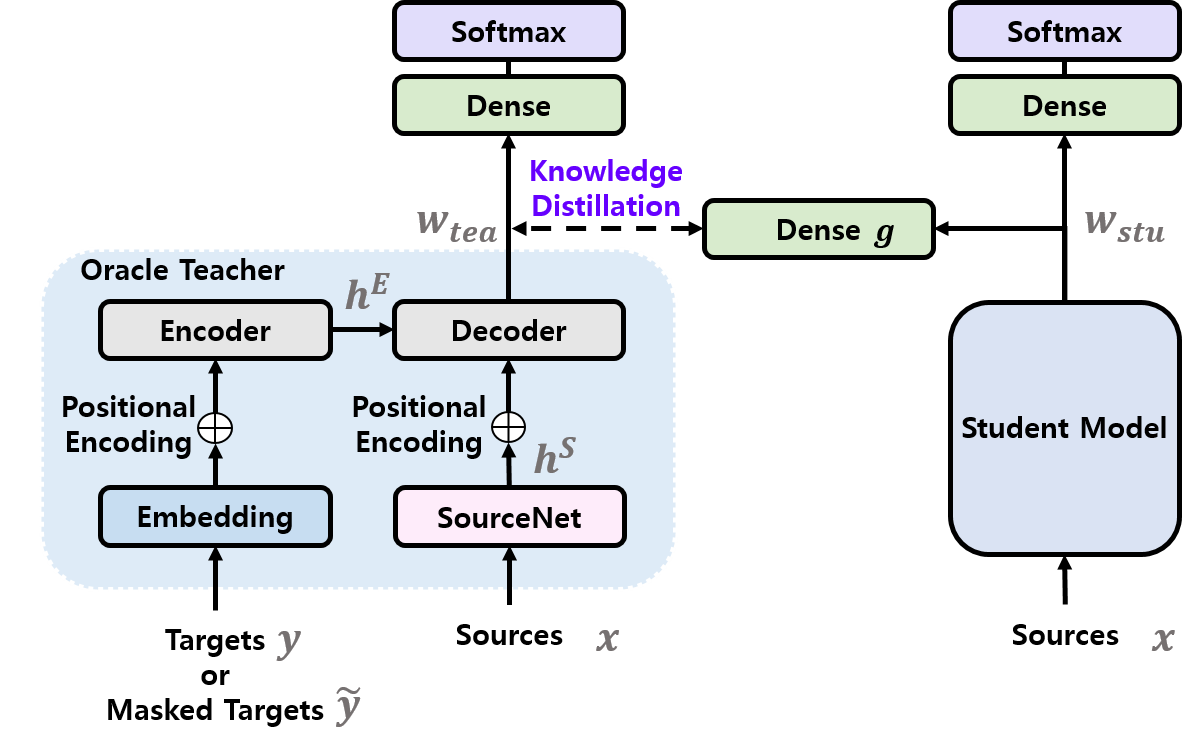}
	\caption{KD procedure with the Oracle Teacher.}
	\label{oracle_kd}
\end{figure}
Directly optimizing the KL-divergence in (\ref{kl}) is intractable because the KL divergence involves the integral that is difficult to calculate.
To sidestep this problem, we can minimize the CE between the softmax outputs of the Oracle Teacher and the student.
However, as reported in previous KD studies \cite{senior-et-al:scheme,takashima-et-al:scheme, takashima-et-al2:scheme}, simply applying the frame-level CE to the CTC-based model can worsen the performance compared to the baseline trained only with the ground truth.
Instead, we adopt FitNets \cite{romero-et-al:scheme} as the basic KD technique, which considers the hidden representation for distillation, and there are two reasons for this choice: (1) In the CTC framework, transferring the hidden representation is much more effective than the softmax-level KD approach \cite{tutornet:scheme}; (2) Recent KD approaches for sequence learning \cite{tutornet:scheme,distillhubert:scheme,fithubert} are based on the Fitnets.

As depicted in Fig. \ref{oracle_kd}, $w_{tea}$ and $w_{stu}$ respectively denote the hidden representation obtained from the last layer of the teacher and student models.
Since usually the hidden layer dimensions of $w_{tea}$ and $w_{stu}$ are different, we apply a fully connected layer $g$ to bridge the dimension mismatch. The process of KD initializes the student by minimizing the distance between hidden representation of the teacher model $w_{tea}$ and student model $w_{stu}$.
The objective for KD is given by
\begin{align}
\label{hidden}
   \mathcal{L}_{KD}(w_{stu}, w_{tea}) =\Vert{w_{tea}-g(w_{stu})}\Vert_2^2.
\end{align}%

\subsection{Model Structure}
\label{oracle_model_structure}
The Oracle Teacher mainly consists of three components: 1) an encoder to encode the target input, 2) a SourceNet to extract the features from the source input data, and 3) a decoder to predict the CTC alignment.
Its architecture follows the encoder-decoder structure of the Transformer \cite{transformer:scheme}, which allows the model to attend to related target information when making a prediction.
Note that the Oracle Teacher is a non-autoregressive model where the look-ahead mask is not included in the decoder.
In Fig. \ref{oracle_arc}, we illustrate the schematics of the Oracle Teacher model which can also be
summarized as follows:
\begin{align}
\label{oracle_align}
  & h^{S}=\text{SourceNet}(x;\theta_{S}), \\
  & h^{E}=\text{Encoder}(y;\theta_{E}), \\
  & P(\pi|x,y) \sim \text{Decoder}(h^{E}, h^{S};\theta_{D}). 
 \end{align}
Compared to the vanilla Transformer, the main architectural difference lies in the cross attention. The encoder takes inputs from the whole target $y$, and its resulting vectors $h^{E}$ are treated as key-value pairs of the cross attention.
In addition, we do not employ a look-ahead mask, which is used in the vanilla Transformer to mask the future tokens, in the multi-head attention layer of the decoder.

\subsubsection{SourceNet}
The SourceNet converts the source input $x$ into high-level representations $h^{S}$.
Since $h^{S}$ serves as query for the decoder, the length of the decoder output has the same length as $h^{S}$.
The architecture of the SourceNet depends on the task we are interested in.
When the main task is ASR. the SourceNet corresponds to an acoustic model part of the conventional ASR model.
In our experiments, we apply CTC-based architecture as the SourceNet, and consequently $|h^{S}|\geq|y|$. 
The Oracle Teacher has the SourceNet with 8 convolutional layers.
In the case of the STR task, we adopt the architecture of the CRNN \cite{crnn:scheme} for the SourceNet.

\subsubsection{Encoder}
In the encoder, we adopt the same structure as the encoder of the original Transformer. 
The self-attention captures dependencies between different positions in $y$ and outputs intermediate representations $h^{E}$.
\subsubsection{Decoder}
The representations $h^{S}$ and $h^{E}$ are fed into the decoder, which follows the architecture of the conventional Transformer decoder.
The self-attention layer, the first attention layer of the decoder, takes the representations $h^{S}$ as the input.
Then, the output serves as queries for the cross attention, whose key-value pairs are the representations $h^{E}$.
The cross attention allows the decoder to look into the relevant target information when producing the prediction.
Note that the look-ahead mask is not included in the decoder.
Different from the autoregressive model that only uses the past output tokens in producing the results, the Oracle Teacher can utilize more global output features when predicting the output.

In the proposed framework, the representation of the source $x$ corresponds to the queries for the cross-attention.
This is because, for KD, the length of the decoder output (= the length of the query $h^S$) should have the same length ($T$) as the student's output, which is determined by the source $x$.
If the decoder output has a different length from that of the student's output, the Oracle Teacher cannot transfer the knowledge to the student.

\section{Experimental Settings}
\subsection{Datasets and Baselines}
\subsubsection{Speech Recogniton}
For ASR, we evaluated the performance of the models on LibriSpeech \cite{librispeech:scheme}. In the training phase, ~``train-clean-100",~``train-clean-360", and~``train-other-500" were used. For evaluation, ~``dev-clean",~``dev-other",~``test-clean", and ~``test-other" were applied. 
We adopted the current high-performing models for the conventional teacher in each task. In the case of ASR, we applied pre-trained Jasper Dense Residual (Jasper DR) \cite{jasper:scheme} with 54 convolutional layers and pre-trained Conformer \cite{conformer} large (Conformer-CTC L) with 18 Conformer layers.
Recent ASR studies \cite{tutornet:scheme, asr1, asr2, asr3} utilized Jasper DR as the baseline.
According to the previous study \cite{jasper:scheme}, Jasper DR produces word error rate (WER) 2.62 \% on dev-clean with strong Transformer-XL \cite{transformer-xl} LM, still SOTA performance on LibriSpeech.
As for the CTC-based student, we used two models: (1) Jasper Mini, composed of 33 depthwise separable 1D convolutional layers, and (2) Conformer small (Conformer-CTC S) consisting of 16 Conformer layers with 176 dimensions.
The Jasper-based Oracle Teacher had the SourceNet with 8 convolutional layers, and both the encoder and decoder consisted of 2 layers.
In the case of the Conformer-based Oracle Teacher, the SourceNet of the Oracle Teacher consisted of 16 Conformer layers with 176 dimensions.
The encoder and decoder of the Oracle Teacher consisted of 1 layer.
When we conducted experiments for the RNN-T framework, we compared the Oracle Teacher with a large Conformer-Transducer model (Conformer-T L), which is the current SOTA for ASR.
Its encoder network consisted of 17 Conformer layers with 512 dimensions.
To extract the knowledge of the Conformer-T L model, we used the pre-trained checkpoint provided by the NeMo \cite{nemo:scheme} toolkit.
The RNN-T student was based on based on a small Conformer model, where the encoder network consisted of 16 layers with 176 dimensions.
For the RNN-T-based Oracle Teacher, the SourceNet had the same architecture with the encoder network of the RNN-T student, consisting of 16 Conformer layers. 
The encoder and decoder of the Oracle Teacher consisted of 1 layer.
In the RNN-T framework, the Oracle Teacher included joint and prediction networks.
The joint and prediction networks had the same architecture as the RNN-T student.
We trained the Oracle Teacher using the hybrid CTC/RNN-T framework, which is the recent RNN-T configuration in NeMo. The additional CTC objective helped achieve stable training for the Transducer-based Oracle Teacher.

\begin{table*}[t]
\centering
\caption{WER (\%) performance comparison across CTC-based ASR models on LibriSpeech. The best result of the student is in bold.}{%
{\fontsize{7.3}{8.76}\selectfont
\label{sr_result}
\begin{tabular}{|c|c|c|c|c|c|c|c|c|c|}
\hline
\multirow{3}{*}{\makecell{ASR baseline\\model}} &\multirow{3}{*}{Params.}& \multicolumn{4}{c|}{w/o LM}  & \multicolumn{4}{c|}{w/ LM}                                       \\ \cline{3-10}
             &        & \multicolumn{2}{c|}{dev} &\multicolumn{2}{c|}{test}& \multicolumn{2}{c|}{dev} &\multicolumn{2}{c|}{test}           \\\cline{3-10}
             &        & clean         & other          & clean         & other          & clean         & other          & clean         & other          \\
\hline
Jasper DR \cite{jasper:scheme}& 333 M & 3.61 & 11.37 & 3.77 & 11.08 & 2.99 & 9.40 & 3.62 & 9.33\\
Jasper Mini & 8 M & 8.66          & 23.28          & 8.85          & 24.26          & 4.78          & 15.14          & 5.15          & 15.77          \\
\hline
\hline
\multirow{3}{*}{Student}&\multirow{3}{*}{Teacher} & \multicolumn{4}{c|}{w/o LM}                                      & \multicolumn{4}{c|}{w/ LM}                                       \\\cline{3-10}
    &        & \multicolumn{2}{c|}{dev}&\multicolumn{2}{c|}{test}& \multicolumn{2}{c|}{dev}&\multicolumn{2}{c|}{test}\\\cline{3-10}
   &         & clean         & other          & clean         & other          & clean         & other          & clean         &other          \\
\hline
\multirow{4}{*}{Jasper Mini}&None             & 8.66          & 23.28          & 8.85          & 24.26          & 4.78          & 15.14          & 5.15          & 15.77          \\
&Jasper DR \cite{jasper:scheme}                & 7.05          & 19.41          & 7.03          & 20.41          & 4.80          & 14.32          & 5.00          & 14.99          \\
&\textbf{Oracle Teacher (ours)}& \textbf{6.64} & \textbf{18.91} & \textbf{6.67} & \textbf{19.82} & \textbf{4.65} & \textbf{14.31} & \textbf{4.90} & \textbf{14.65} \\
&Oracle Teacher w/o target   & 7.22          & 20.39              & 7.32             & 21.10              & 4.72            & 14.67              & 4.91             & 15.15       \\     
\hline
\end{tabular}}}
\end{table*}

\subsubsection{Scene Text Recognition}
We evaluated STR models on seven benchmark datasets\footnote{We applied the datasets used in the comparative study conducted by Baek \textit{et al.} \cite{clova:scheme}.}: Street View Text (SVT) \cite{svt:scheme}, SVT Perspective (SVTP) \cite{svtp:scheme}, IIIT5K-Words (IIIT) \cite{iiit:scheme}, CUTE80 (CT) \cite{cute:scheme}, ICDAR03 (IC03) \cite{ic03:scheme}, ICDAR13 (IC13) \cite{ic13:scheme}, and ICDAR15 (IC15) \cite{ic15:scheme}.
For validation, IC13, IC15, IIIT, and SVT were applied. 
As training datasets, we used the two most popular datasets: MJSynth \cite{mj:scheme} and SynthText \cite{st:scheme}. 
We adopted Rosetta \cite{rosetta:scheme} and STAR-Net \cite{starnet:scheme}, considered as the benchmarking SOTA models in recent researches \cite{textocr, ocr2}.
In the case of the student model, CRNN \cite{crnn:scheme} was adopted with a thin-plate spline (TPS), which normalizes curved and perspective texts into a standardized view.
The SourceNet followed the TPS-CRNN structure, and both the encoder and decoder used 1 layer.

\subsection{Implementation Details}
\subsubsection{Speech Recognition}
For the LibriSpeech dataset, We used the OpenSeq2Seq \cite{openseq2seq} and NeMo \cite{nemo:scheme} toolkits for the implementation. 
In the case of Jasper-based ASR models, they were based on character-level CTC models. 
The character set had 29 labels. In the case of Jasper DR, we used the pre-trained model provided by the OpenSeq2Seq toolkit. The student model was run on three Titan V GPUs, each with 12GB of memory.
We used a NovoGrad optimizer \cite{ginsburg-et-al:scheme} whose initial learning rate started from 0.02 with a weight decay of 0.001. 
When applying the Conformer-based student model, we used byte-pair encoding (BPE) \cite{bpe:scheme} tokens as the output units. It was run on four Quadro RTX 8000 GPUs.
We employed a AdamW optimizer with the initial learning rate 5.0.
For KD, the Japser-based student was initially trained with FitNets \cite{romero-et-al:scheme} loss for 5 epochs.
After initialization, 50 epochs were spent on CTC training for the Jasper-based student model
In the case of RNN-T, the student was trained with FitNets for 10 epochs.
After that, it was trained with the RNN-T training for 100 epochs.
In addition, we trained the Jasper-based Oracle Teacher for 30 epochs on a single Titan V GPU using Noam learning rate scheduler with 4000 steps of warmup and a learning rate of 1.5. 
When applying beam-search decoding with language model (LM), we used KenLM \cite{Heafield2011KenLMFA} for 4-gram LM, where the LM weight, the word insertion weight, and the beam width were experimentally set to 2.0, 1.5, and 512, respectively.
The Oracle Teacher for Conformer-CTC model was trained for 15 epochs on four Quadro RTX 8000 GPUs, adopting the AdamW optimizer with the initial learning rate 5.0.
In the case of RNN-T-based Oracle Teacher, it was trained for 25 epochs on four Quadro RTX 8000 GPUs.
It also adopted the AdamW optimizer with the initial learning rate 5.0.

For the Mandarin ASR dataset, the character set had a total of 5207 labels. Pre-trained Jasper DR, which was used as the conventional teacher, was provided by the NeMo \cite{nemo:scheme} toolkit. The student was trained in an identical way to LibriSpeech, but the initial learning rate was set to 0.005.
Instead of WER, we measured the character error rate (CER) since a single character often represents a word for Mandarin. 

\subsubsection{Scene Text Recognition}
When training the STR models, our experiments were conducted using the official implementation provided by Baek \textit{et al.}\footnote{https://github.com/clovaai/deep-text-recognition-benchmark} \cite{clova:scheme}. 
STR models were based on the character-level CTC models.
The character set had a total of 37 labels.
All STR models, including the Oracle Teacher, were trained for 300k iterations on a single Titan V GPU (12GB) in the CTC framework.
We employed the AdaDelta optimizer \cite{adadelta:scheme} with a decay rate of 0.95, and the initial learning rate was 1.0.
In FitNets \cite{romero-et-al:scheme} training, we trained 300k iterations for the student.

\begin{table*}[t]
\centering
\caption{CER (\%) on AISHELL-2 when greedy decoding was applied. The best result of the student is in bold.}{%
{\fontsize{7.3}{8.76}\selectfont
\label{sr_result_aishell}
\begin{tabular}{|c|c|c|c|c|c|c|c|}
\hline
\multirow{2}{*}{\makecell{ASR baseline\\model} } & \multirow{2}{*}{Params.} &  \multicolumn{3}{c|}{dev} & \multicolumn{3}{c|}{test}\\ \cline{3-8}
  &          & iOS & Android & Mic & iOS & Android & Mic\\
\hline
Jasper DR \cite{jasper:scheme} & 338 M & 9.69 & 11.48 & 12.23 & 9.37 & 10.84 & 11.84\\
Jasper Mini& 14 M & 11.77 & 14.23 & 15.03 & 11.38 & 12.71 & 14.27 \\
\hline
\hline
\multirow{2}{*}{Student} & \multirow{2}{*}{Teacher} &  \multicolumn{3}{c|}{dev} & \multicolumn{3}{c|}{test}\\ \cline{3-8}
  &          & iOS & Android & Mic & iOS & Android & Mic\\
\hline
\multirow{4}{*}{Jasper Mini} &None & 11.77 & 14.23 & 15.03 & 11.38 & 12.71 & 14.27\\
&Jasper DR \cite{jasper:scheme}                 & 10.70&12.78&13.66&10.12&11.31&12.60 \\
&\textbf{Oracle Teacher (ours)}& \textbf{9.74}&\textbf{11.49}&\textbf{12.31}&\textbf{9.27}&\textbf{10.36}&\textbf{11.99}\\
&Oracle Teacher w/o target   & 10.45&12.42&13.13&9.76&10.92&12.19\\      
\hline
\end{tabular}}
}
\end{table*}

\section{Experimental Results}
In the subsequent part of this paper, $A$ $\rightarrow$ $B$ means that teacher model $A$ transfers knowledge to the student model $B$.
As mentioned in Section \ref{oracle_kd_section}, we employed FitNets \cite{romero-et-al:scheme} as the basic KD technique.

\subsection{Main Results: Performance Comparison}
Since the Oracle Teacher is the teacher model for KD, not the baseline model performing the learning task, the evaluation results of the Oracle Teacher itself are not described in this section. Note that the model size and the performance of the Oracle Teacher will be additionally discussed in Section \ref{computational cost section} and \ref{per_oracle}.
\begin{table*}[t]
\centering
\caption{Performance of CTC-based STR models. The best result of the student is in bold.}{%
{\fontsize{7.3}{8.76}\selectfont
\centering
\begin{tabular}{|c|c|c|c|c|c|c|c|c|c|c|c|c|}
\hline
\makecell{STR baseline\\model} & Params. & \makecell{IIIT\\3000}     & SVT             & \makecell{IC03\\860}         & \makecell{IC13\\857}        &\makecell{IC13\\1015}       & \makecell{IC15\\1811}       & \makecell{IC15\\2077}       & SVTP            & CT          & \makecell{Total\\accuracy}  \\
\hline
Rosetta \cite{rosetta:scheme} & 46 M & 85.53 & 84.85 &94.19  &91.95 &90.74 &73.22 &70.55 &76.12 &68.99 &82.45 \\
Star-Net \cite{starnet:scheme} & 49 M & 85.50 & 85.47 & 93.84  &92.77 & 91.92 &72.50 &69.77 &73.80 &70.38 &82.24\\
CRNN \cite{crnn:scheme}  & 10 M  & 83.87          & 80.37          & 93.02            & 90.43          & 89.46          & 70.07          & 67.53          & 72.09          & 65.51          & 80.10          \\
 \hline
 \hline
  Student&Teacher                   & \makecell{IIIT\\3000}     & SVT             & \makecell{IC03\\860}            & \makecell{IC13\\857}        &\makecell{IC13\\1015}       & \makecell{IC15\\1811}       & \makecell{IC15\\2077}       & SVTP            & CT          & \makecell{Total\\accuracy}  \\
\hline
\multirow{5}{*}{CRNN \cite{crnn:scheme}} & None    & 83.87          & 80.37          & 93.02              & 90.43          & 89.46          & 70.07          & 67.53          & 72.09          & 65.51          & 80.10          \\
&Rosetta \cite{rosetta:scheme} &   84.70          & 83.46          & 92.91                 & 91.02          & 90.15          & 71.89          & 69.20          & 71.16          & 65.85          & 81.04          \\
&Star-Net \cite{starnet:scheme} &   85.20          & 84.39          & 93.49         & \textbf{91.60} & \textbf{90.74} & 72.45          & 69.77          & 72.25          & 70.04          & 81.77          \\
&\textbf{Oracle Teacher (ours)}& \textbf{85.77} & \textbf{84.54} & \textbf{93.61}         & 91.48          & 90.54          & \textbf{73.11} & \textbf{70.40} & \textbf{74.26} & \textbf{70.38} & \textbf{82.21} \\
&Oracle Teacher w/o target& 85.40 & 82.84 & 93.02 & 90.78 & 89.75 & 71.73 & 69.04 & 72.71 & 68.99 & 81.30 \\
\hline
\end{tabular}%
\label{str_results}
}}
\end{table*}

\subsubsection{Speech Recognition} 
The results for LibriSpeech are shown in Table \ref{sr_result}.
We measured WER to quantify the performance.
The best performance was achieved when training the student with the Oracle Teacher.
In addition, to further check the effectiveness of the target input $y$, which is used as the additional input of the Oracle Teacher, we applied an incomplete Oracle Teacher model, called Oracle Teacher w/o target.
In Oracle Teacher w/o target, zero arrays were treated as the additional input instead of the target input $y$ during training and inference phases.
Since the Oracle Teacher w/o target only consumed the source input, its architecture was similar to that of the conventional CTC model.
When applying the Oracle Teacher w/o target, the distilled student achieved improvement over the baseline student, which indicates that the knowledge of the SourceNet contributed to improving the performance of the student.
However, their performances were worse than the Oracle Teacher $\rightarrow$ Jasper Mini in all configurations, implying that the oracle guidance helped the Oracle Teacher extract a more supportive knowledge for the student.

As presented in Table \ref{sr_result_aishell}, we can confirm that the Oracle Teacher still works well with KD on the Mandarin dataset.
Interestingly, when the Oracle Teacher was applied, the distilled student (CER: 9.74 \% on dev-iOS) performed similarly to the pre-trained Jasper DR (CER: 9.69 \% on dev-iOS), notwithstanding its smaller parameter size (14 M parameters) than Jasper DR (333 M parameters). In some cases, including test-iOS and test-Android, the student distilled from the Oracle teacher outperformed the Jasper DR teacher.
When transferring the knowledge from the Oracle Teacher w/o target, the results show that Oracle Teacher w/o target $\rightarrow$ Jasper Mini performed better than Jasper DR $\rightarrow$ Jasper Mini.
It indicates that, even without the additional target information, the student can benefit from the knowledge of the SourceNet.
However, our best performance was achieved when applying the Oracle Teacher as the teacher model.

\subsubsection{Scene Text Recognition} 
For the STR task, we used accuracy, the success rate of word predictions per image, as a performance metric. 
As reported in Table \ref{str_results}, the student distilled from the Oracle Teacher showed better performance than those distilled from other teachers, and its total accuracy (82.21 \%) was almost similar to that of the conventional teacher Star-Net (82.24 \%) while having much fewer parameters (10 M parameters).
On IC13 datasets, the performances of Star-Net $\rightarrow$ CRNN were slightly better than those of Oracle Teacher $\rightarrow$ CRNN.
However, the differences were negligible since Oracle Teacher $\rightarrow$ CRNN showed better improvements in most cases, including the total accuracy.
Oracle Teacher w/o target $\rightarrow$ CRNN performed better than Rosetta $\rightarrow$ CRNN in some cases. 
It means that, even without using the additional target input, the student can benefit from the knowledge of the SourceNet. 
However, the distilled student from the Oracle Teacher w/o target had worse achievements than Star-Net $\rightarrow$ CRNN and Oracle Teacher $\rightarrow$ CRNN, indicating that the target input played an important role in the effectiveness of the Oracle Teacher.

\begin{figure*}[t]
\centering
	\includegraphics[height=10cm]{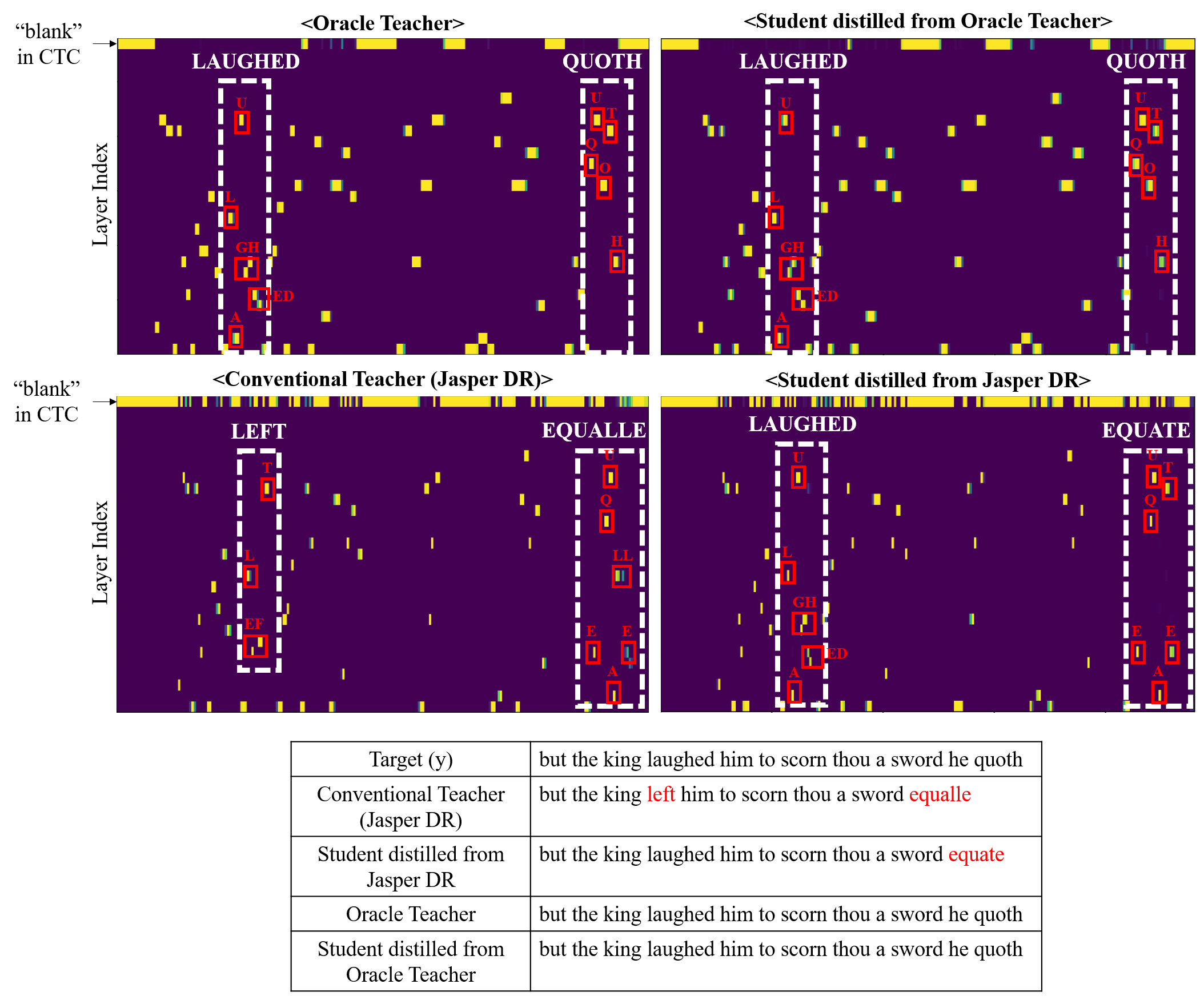}
	\caption{Frame-wise label probability example on LibriSpeech test-other dataset. Conventional teacher denotes the Jasper DR model. The x-axis refers to acoustic frames, and the y-axis refers to the character labels. The last label index represents the ``blank" label in the CTC framework.}
	\label{case_asr}
\end{figure*}

\subsection{Case Study: The Effect of Target Input}
\label{case_sec}

To validate why the proposed method could result in better KD performance than the conventional teacher, we conducted a case study for ASR on LibriSpeech test-other dataset.
By comparing predictions between the conventional teacher and the Oracle Teacher, we verified the effect of using the target information and the behaviour of the Oracle Teacher.

In Fig. \ref{case_asr}, we visualized the softmax prediction (CTC alignment) of the conventional teacher and the Oracle Teacher.
The x-axis refers to acoustic frames, and the y-axis refers to the character labels.
As displayed in Fig. \ref{case_asr}, the conventional teacher converted a given speech into ``but the king left him to scorn thou a sword equalle" and made erroneous predictions with ``left" and ``equalle".
When conditioning on the speech voice only, it is hard to distinguish ``left"/``laughed" and ``equalle"/``he quoth".
However, the Oracle Teacher gave accurate CTC alignment by utilizing the additional target (text) information, implying that a more optimal problem-solving could be derived by referring to both source (speech) and target (text) information.

Also, the distilled student properly learned the behavior of the Oracle Teacher. 
As shown in Fig. \ref{case_asr}, the student distilled from the conventional teacher could not distinguish ``equate”/``he quoth”, indicating that the knowledge of the conventional teacher led to a sub-optimal solution for training the student.
In contrast, the distilled student from the Oracle Teacher produced an accurate prediction. This means that the student could effectively benefit from the better alignment of the Oracle Teacher. Since the Oracle Teacher produced more accurate alignment by using the additional text oracle information, it could transfer a more optimal knowledge to the student.

In CTC, the blank label relieves the network from making label predictions at a frame when it is uncertain \cite{graves-et-al:scheme, uncertain1, uncertain2}. Interestingly, as shown in Fig. \ref{case_asr}, most frames of the conventional teacher had the highest probability for the ``blank" token, meaning that the model was uncertain about the corresponding acoustic regions. In contrast, the Oracle Teacher had fewer frames that were predicted as ``blank" token. Based on the linguistic information from the text input, multiple frames of the Oracle Teacher were more likely to be predicted as non-blank character labels rather than the ``blank" during the active speech periods.
Thus, the Oracle Teacher was likely to have less uncertainty in many acoustic regions while containing much more information about non-blank labels. Since the Oracle Teacher could achieve an accurate alignment and less uncertainty on most frames, the knowledge of the Oracle Teacher was more helpful for training the student.

\begin{figure}[t]
\centering
	\includegraphics[height=2cm]{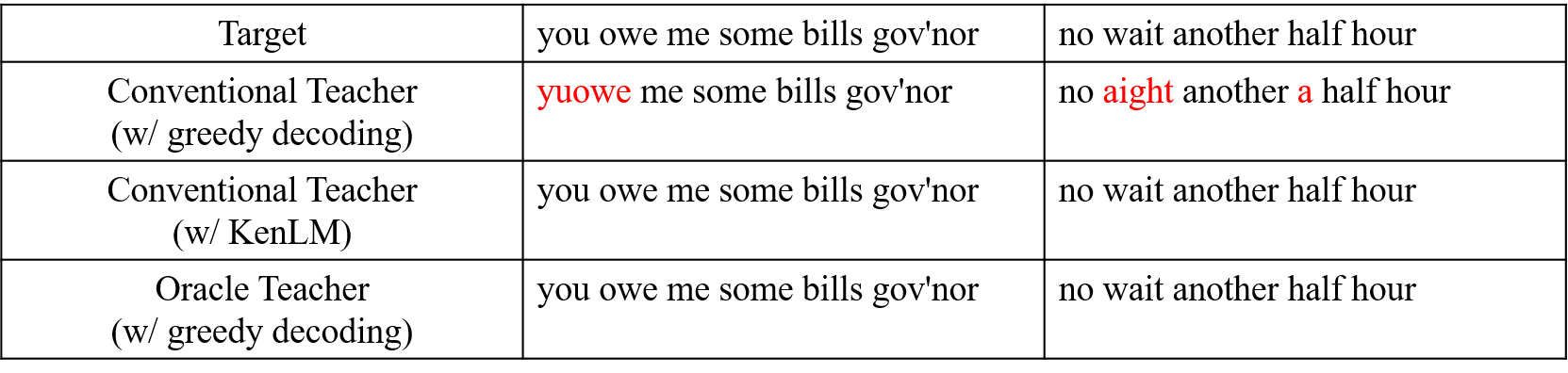}
	\caption{Recognition example on LibriSpeech test-clean dataset.}
	\label{case_asr2}
\end{figure}

\begin{figure*}[t]
\centering
	\includegraphics[height=7cm]{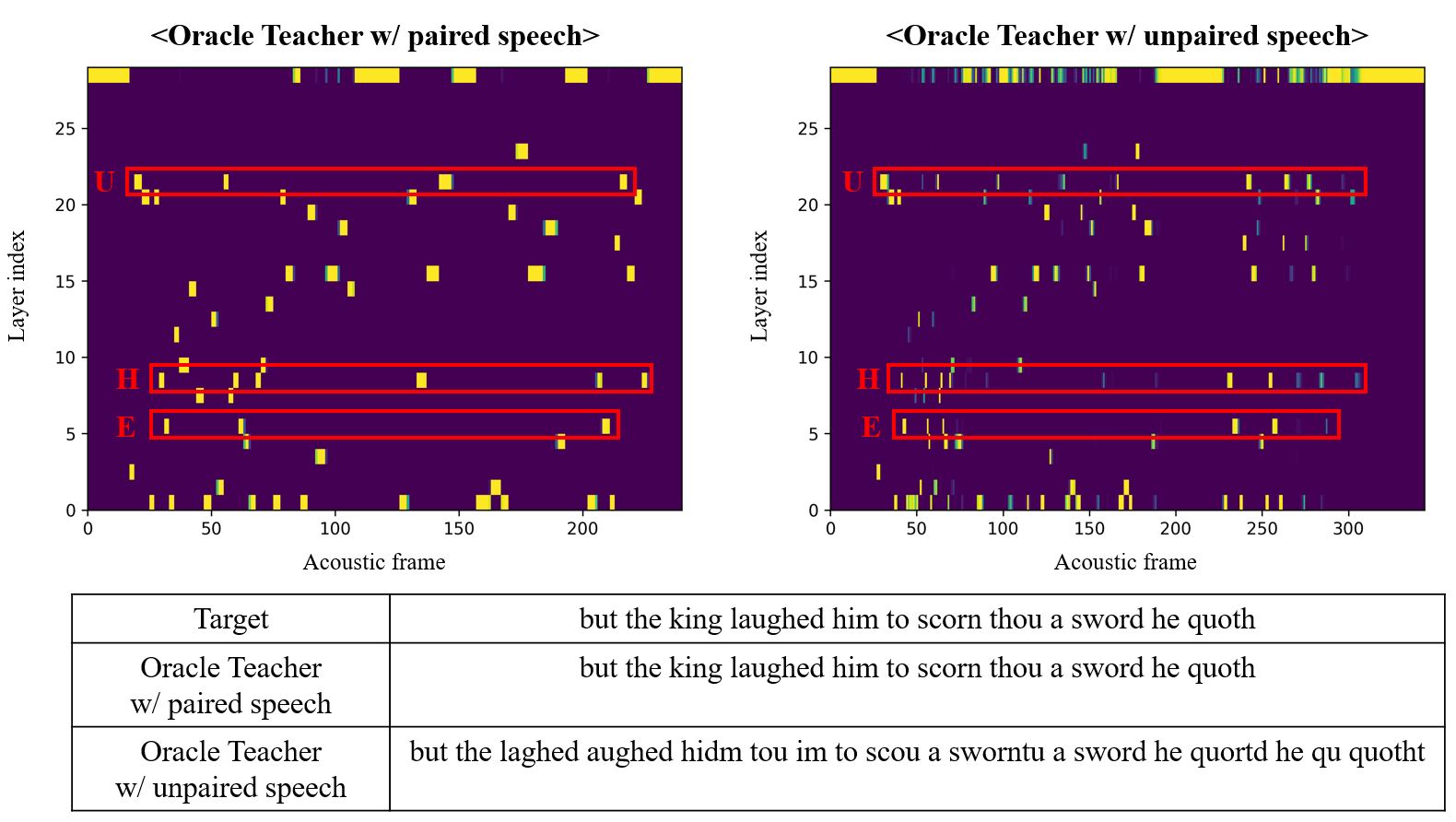}
	\caption{Frame-wise label probability example on LibriSpeech test-other dataset. The x-axis refers to acoustic frames, and the y-axis refers to the character labels. The last label index represents the ``blank" label in the CTC framework.}
	\label{unpaired}
\end{figure*}

\begin{figure}[t]
\centering
	\includegraphics[height=10cm]{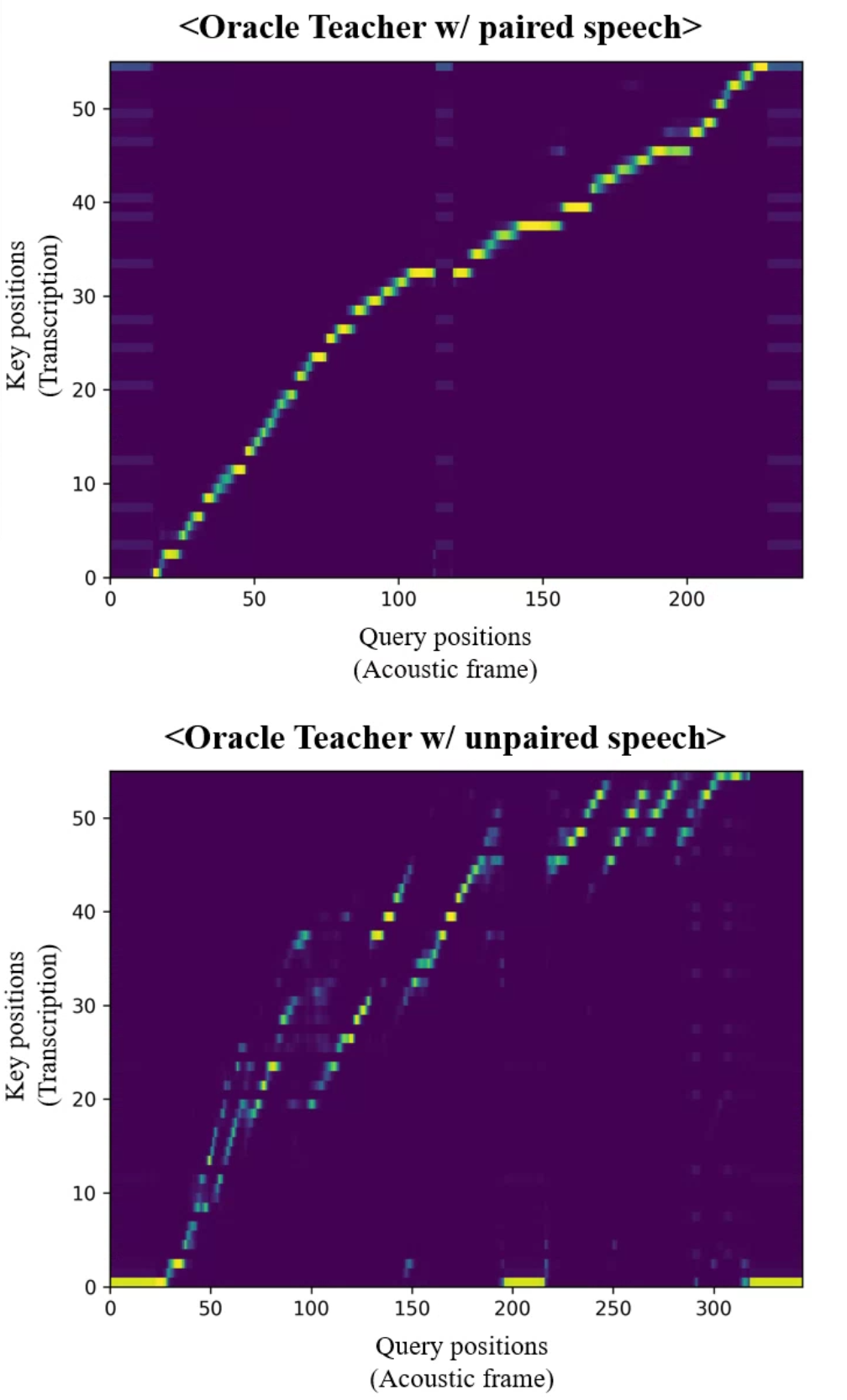}
	\caption{Visualized of the attention weight of the decoder.}
	\label{unpaired_attention}
\end{figure}

In addition, we compared the ASR predictions of the conventional teacher and the Oracle Teacher, as shown in Fig. \ref{case_asr2}.
In Fig. \ref{case_asr2}, the conventional teacher made erroneous predictions with ``yuowe", ``aight", and ``a" using the greedy decoding.
When considering the acoustic (speech) feature only, it is challenging to distinguish some words, such as ``you owe"/``yuowe" and ``wait"/``aight".
The conventional teacher generated an accurate prediction when decoding with the external KenLM that provided additional linguistic information.
However, the proposed Oracle Teacher could produce correct ASR prediction without using the external LM.
This is because the Oracle Teacher leveraged both the source input (speech) and the output label (text) as the teacher model's input.
Unlike the conventional teacher that only considered acoustic features, the Oracle Teacher performed a fusion of acoustic (speech) and linguistic (text) features when generating the prediction.
Since unifying acoustic and linguistic representation learning generally enhances the performance of the speech processing \cite{acours_lin1,acours_lin2,acours_lin3,acours_lin4,acours_lin5}, the Oracle Teacher that considered linguistic information could estimate better CTC prediction, and also its representation was a more supportive knowledge for the ASR student.

\subsection{Case Study: Linguistic Information from Encoder} 
\label{linguistic}
To further check the behavior of the proposed teacher, we fed unpaired speech and text inputs to the pre-trained Oracle Teacher, allowing us to explore the model's behavior regardless of the speech input.
Note that the Oracle Teacher was already pre-trained with paired speech and text inputs.
In this experiment, ``unpaired speech" means a speech that was unrelated to the transcription.
For unpaired speech, we randomly selected an utterance from the LibriSpeech when the transcription was ``but the king laughed him to scorn thou a sword he quoth".
In Fig. \ref{unpaired}, we visualized the recognition examples and softmax predictions of the Oracle Teacher with paired speech (original Oracle Teacher) and the Oracle Teacher with unpaired speech. 
Since the paired and unpaired speeches had different lengths, the acoustic frames were different.
The results showed that the encoder sufficiently captured linguistic information from the ground truth. When using the unpaired speech input, even though some information related to the word ``king" might be lost, the prediction still included information about other characters.

However, we observed that the encoder alone could not assign suitable text information for each frame. As illustrated in Fig. \ref{unpaired}, the Oracle Teacher with unpaired speech produced incorrect prediction with many unnecessary repetitions, such as ``E”, ``H”, and ``U”.
Also, in Fig. \ref{unpaired_attention}, we visualize the cross attention scores of the Oracle Teacher, where the x-axis refers to acoustic frames and the y-axis refers to the transcription.
When using unpaired speech, the attention scores were incorrectly computed for the alignment between speech and text. 
In contrast, as shown in Fig. \ref{unpaired_attention}, the Oracle Teacher with paired speech correctly computed the alignment along with the acoustic frames (query).
This means that aligning linguistic information to each frame depended on the acoustic information from SourceNet.

In the sequence-to-sequence problem, learning the alignment between the source (speech) and target (text) is important but difficult due to the length mismatch. To achieve a good WER score, conventional ASR models are trained to learn 1) the linguistic information from speech input and 2) the alignment between speech and text. However, the Oracle Teacher already had sufficient linguistic information from the text input. Based on the provided linguistic information, it could also easily learn the alignment between speech and text. Since both linguistic and alignment information existed perfectly in the proposed framework, the Oracle Teacher could provide the optimal knowledge to the student model while having fewer parameters.

Additionally, we confirmed that the proposed teacher did not just mimic the text input. If the Oracle Teacher simply copied the ground-truth as the output, then it produced an accurate prediction regardless of the speech input. However, when we fed the model with unpaired speech, it produced incorrect predictions, while the prediction of the original Oracle Teacher was accurate.

\begin{table}[t]
\centering
{\fontsize{7.3}{8.76}\selectfont
\centering
\caption{WER (\%) performance comparison across CTC-based ASR models on LibriSpeech. The best result of the student is in bold.}
\label{asr_app}
\begin{tabular}{|c|c|c|cc|cc|}
\hline
\multirow{2}{*}{Teacher} & \multirow{2}{*}{Student}     & \multirow{2}{*}{KD method} & \multicolumn{2}{c|}{dev}                            & \multicolumn{2}{c|}{test}                           \\ \cline{4-7} 
                         &                              &                            & \multicolumn{1}{c|}{clean}         & other          & \multicolumn{1}{c|}{clean}         & other          \\ \hline
None                     & \multirow{7}{*}{\makecell{Jasper\\Mini}}                  & None                       & \multicolumn{1}{c|}{8.66}          & 23.28          & \multicolumn{1}{c|}{8.85}          & 24.26          \\ \cline{1-1} \cline{3-7} 
Jasper DR                &  & \multirow{2}{*}{RKD \cite{tutornet:scheme}}                       & \multicolumn{1}{c|}{6.74}          & 19.27          & \multicolumn{1}{c|}{6.77}          & 19.78          \\ \cline{1-1} \cline{4-7} 
Oracle Teacher           &                              &                       & \multicolumn{1}{c|}{\textbf{6.44}} & \textbf{18.36} & \multicolumn{1}{c|}{\textbf{6.43}} & \textbf{18.97} \\ \cline{1-1} \cline{3-7} 
Jasper DR                &                              & \multirow{2}{*}{SKD \cite{tutornet:scheme}}                          & \multicolumn{1}{c|}{7.64}          & 21.36          & \multicolumn{1}{c|}{7.81}          & 22.41          \\ \cline{1-1} \cline{4-7} 
Oracle Teacher           &                              &                       & \multicolumn{1}{c|}{7.57}          & 21.20          & \multicolumn{1}{c|}{7.71}          & 21.71          \\ \cline{1-1} \cline{3-7} 
Jasper DR                &                              & \multirow{2}{*}{FitNets \cite{romero-et-al:scheme}}                      & \multicolumn{1}{c|}{7.05}          & 19.41          & \multicolumn{1}{c|}{7.03}          & 20.41          \\ \cline{1-1} \cline{4-7} 
Oracle Teacher           &                              &                    & \multicolumn{1}{c|}{6.64}          & 18.91          & \multicolumn{1}{c|}{6.67}          & 19.82          \\ \hline
\end{tabular}}
\end{table}

\begin{table}[t]
\centering
\caption{WER (\%) performance comparison on LibriSpeech with greedy decoding.}
{\fontsize{7.3}{8.76}\selectfont
\begin{tabular}{|c|c|cc|cc|}
\hline
\multirow{2}{*}{\begin{tabular}[c]{@{}c@{}}ASR baseline\\ model\end{tabular}} & \multirow{2}{*}{Params.} & \multicolumn{2}{c|}{dev} & \multicolumn{2}{c|}{test} \\ \cline{3-6} 
 &  & \multicolumn{1}{c|}{clean} & other & \multicolumn{1}{c|}{clean} & other \\ \hline
Conformer-CTC L & 122 M & \multicolumn{1}{c|}{2.47} & 6.03 & \multicolumn{1}{c|}{2.78} & 6.18 \\
Conformer-CTC S & 13 M & \multicolumn{1}{c|}{4.63} & 12.21 & \multicolumn{1}{c|}{4.87} & 12.05 \\ \hline \hline
\multirow{2}{*}{Student} & \multirow{2}{*}{Teacher} & \multicolumn{2}{c|}{dev} & \multicolumn{2}{c|}{test} \\ \cline{3-6} 
 &  & \multicolumn{1}{c|}{clean} & other & \multicolumn{1}{c|}{clean} & other \\ \hline
\multirow{3}{*}{Conformer-CTC S} & None & \multicolumn{1}{c|}{4.63} & 12.21 & \multicolumn{1}{c|}{4.87} & 12.05 \\
 & Conformer-CTC L & \multicolumn{1}{c|}{4.20} & 10.86 & \multicolumn{1}{c|}{4.20} & 10.90 \\
 & Oracle Teacher & \multicolumn{1}{c|}{4.02} & 10.77 & \multicolumn{1}{c|}{4.10} & 10.39 \\ \hline
\end{tabular}}
\label{conformer-ctc}
\end{table}

\begin{table*}[t]
\centering
\caption{Computational resource consumption comparison across teacher models.}{%
{\fontsize{7.3}{8.76}\selectfont
\label{gpu}
\begin{tabular}{|c|c|c|c|c|c|c|c|}
\hline
Task &Training dataset& Teacher model & Params.& GPU & Batch & Times & Epochs \\
\hline
\multirow{2}{*}{ASR}&\multirow{2}{*}{LibriSpeech}&Jasper DR \cite{jasper:scheme}&333 M&8 * 32GB& 256 & - &400 epochs\\
&&\textbf{Oracle Teacher (ours)}&33 M&1 * 12GB& 64 & 22 h &30 epochs\\
\hline
\multirow{2}{*}{ASR}&\multirow{2}{*}{LibriSpeech}&Conformer-CTC L &122 M&128 * 50GB& 2048 & - &1000 epochs\\
&&\textbf{Oracle Teacher (ours)}&14 M&4 * 45GB& 224 & 8 h &15 epochs\\
\hline
\multirow{2}{*}{ASR}&\multirow{2}{*}{AISHELL-2}&Jasper DR\cite{jasper:scheme} &338 M&8 * 32GB& 128 & - &50 epochs\\
&&\textbf{Oracle Teacher (ours)}&34 M&1 * 12GB& 64 & 118 h &30 epochs\\
\hline
&&Star-Net \cite{starnet:scheme} &49 M&1 * 12GB& 192 &27 h & 300k iter.\\
STR&MJSynth + SynthText&Rosetta \cite{rosetta:scheme}&46 M&1 * 12GB& 192 &27 h& 300k iter.\\
&&\textbf{Oracle Teacher (ours)}& 12 M&1 * 12GB& 192 &10 h& 300k iter.\\
\hline
\end{tabular}%
}}
\end{table*}

\subsection{Performance Comparison with Other KD Methods}

In the previous results, we applied Fitnets \cite{romero-et-al:scheme} as the basic KD method. To further validate the effectiveness of the Oracle Teacher, we used other KD methods for performance comparison.

Firstly, we applied RKD \cite{tutornet:scheme} as the KD method, a recent KD approach in ASR task. It transfers the representation-level knowledge by considering a frame weighting, reflecting which frames were important for KD. 
In addition to the RKD, we adopted SKD \cite{tutornet:scheme} for KD, which effectively transfers the softmax-level knowledge in the CTC framework.
From the results in Table \ref{asr_app}, it is confirmed that Oracle Teacher was more supportive than the conventional Jasper DR teacher in all configurations.
Also, we verified that RKD achieved better improvements than other KD methods, including FitNets and SKD. The best performance was achieved when using the Oracle Teacher with the RKD. We can observe the consistent performance gain of the Oracle Teacher over the conventional teacher for various KD methods.

To further improve the WER performance, we adopted Conformer \cite{conformer}-CTC as the baseline.
In this setting, the SourceNet of the Oracle Teacher consisted of 16 Conformer layers with 176 dimensions.
Also, both RKD and SKD were applied as KD methods.
The Conformer-based student was trained with RKD loss for 10 epochs, and 100 epochs were spent on CTC training with SKD.
Table \ref{conformer-ctc} gives the WER results on LibriSpeech with greedy decoding. From the results, we observe that the student (Conformer-CTC S) distilled from the Oracle Teacher achieved WER 10.39 \% (RERR: 13.78 \%) on test-other with greedy decoding, while the student distilled from a large Conformer-CTC model (Conformer-CTC L) produced WER 10.90 \% (RERR: 9.54 \%). This indicates that the Oracle Teacher was more supportive in distilling the knowledge compared to the Conformer-CTC L.
Considering that the Conformer-CTC L is the current SOTA ASR model, the KD using the proposed framework was quite effective and efficient.

\subsection{Computational Cost Comparison} 
\label{computational cost section}
In addition to the previous experiments, we proceeded to verify the computational efficiency of the proposed teacher model. Computational resource consumption compared to the conventional teacher models are shown in Table \ref{gpu}.
\subsubsection{Speech Recognition}
Since it is difficult to reproduce the reported WER results of Jasper DR \cite{jasper:scheme} without a large number of resources, we used the checkpoint for LibriSpeech, provided by the OpenSeq2Seq \cite{openseq2seq} toolkit. 
For LibriSpeech, the pre-trained Jasper DR model required eight 32GB GPUs for 400 epochs with a batch size of 256. Its training time had not been reported previously, either in the paper of Li \textit{et al.} \cite{jasper:scheme} or the toolkit.
In the case of the Oracle Teacher, we trained the model for 30 epochs on a single 12GB GPU, which took about 22 hours ($\approx$ 1 day) to finish the training.
Considering that the reported training of the Quartznet \cite{quartznet:scheme}, which is more computationally efficient than Jasper DR, for 400 epochs took 122 hours ($\approx$ 5 days) with eight 32GB GPUs with a batch size of 256, the Oracle Teacher dramatically reduced the computational cost of the teacher model.
Note that the SourceNet consisted of 8 convolutional layers and was not based on a large Jasper DR architecture.
Since the Oracle Teacher used a small SourceNet, the number of parameters in the Oracle Teahcer was about 33M. Also, we trained the Oracle Teacher from the scratch and did not use the pre-trained model parameters.

In the case of Conformer-CTC, the training of Conformer-CTC L required 128 GPUs for 1000 epochs with a batch size of 2048.
Compared to the conventional Conformer-CTC teacher model, our framework dramatically reduced the computational cost to train the teacher model.
\subsubsection{Scene Text Recognition}
As presented in Table \ref{gpu}, the training of Star-Net \cite{starnet:scheme} took about 27 hours on a single Titan V GPU (12GB) with a total batch size of 192, and the training of Rosetta \cite{rosetta:scheme} required about 27 hours.
Compared to the two conventional models, the training of Oracle Teacher consumed much less training time (10 hours) with the same computational resource.

\begin{table}[t]
\centering
\caption{Comparison of WER (\%) on LibriSpeech. Teacher model is Oracle Teacher, and we varied the SourceNet's size.}
\fontsize{7.3}{8.76}\selectfont
\label{sourcenet_size}
\begin{tabular}{|c|c|cc|cc|}
\hline
\multirow{2}{*}{\begin{tabular}[c]{@{}c@{}}Oracle Teacher's \\ params.\end{tabular}} & \multirow{2}{*}{Student} & \multicolumn{2}{c|}{dev} & \multicolumn{2}{c|}{test} \\ \cline{3-6} 
 &  & \multicolumn{1}{c|}{clean} & other & \multicolumn{1}{c|}{clean} & other \\ \hline
33 M & \multirow{2}{*}{Jasper Mini} & \multicolumn{1}{c|}{6.64} & 18.91 & \multicolumn{1}{c|}{6.67} & 19.82 \\ \cline{1-1} \cline{3-6} 
76 M &  & \multicolumn{1}{c|}{6.97} & 19.80 & \multicolumn{1}{c|}{6.97} & 20.17 \\ \hline
\end{tabular}
\end{table}

\subsection{The Effect of SourceNet Size}
The proposed framework yielded considerable performance improvement while achieving better computational efficiency.
To further check the effect of the SourceNet of the Oracle Teacher, we changed the SourceNet's model size.
For ease of comparison, we let Oracle Teacher-S (about 33M) and Oracle Teacher-L (about 76M) denote the original Oracle Teacher and bigger Oracle Teacher, respectively. 
The only difference between the Oracle Teacher-S and Oracle Teacher-L was the size of the SourceNet. 
As shown in Table \ref{sourcenet_size}, interestingly, the student distilled from the Oracle Teacher-S performed better than another. 
In the KD framework, the size (capacity) gap between the teacher and student is important. When the size gap is small, KD could be more effective \cite{gap_kd,gap_kd2}. 
Since SourceNet’s architecture of the Oracle Teacher-S had a similar size to that of the student, the Oracle Teacher-S was more supportive in training the student compared to the Oracle Teacher-L.

\subsection{Analysis}
From the previous experimental results, we validate the superiority of the proposed teacher model.
Therefore, it is necessary to test if the model has been correctly trained.
\subsubsection{Visualization of Cross Attention}
\label{visual_cross}

\begin{figure}[t]
\centering
\begin{subfigure}[Oracle Teacher]
	{
		\includegraphics[height=2.2cm]{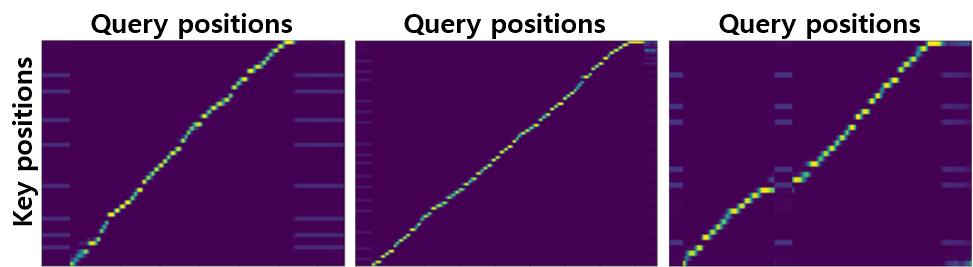}
		\label{attention1}
	}
\end{subfigure}
\begin{subfigure}[Oracle Teacher w/o source]
	{
		\includegraphics[height=2.2cm]{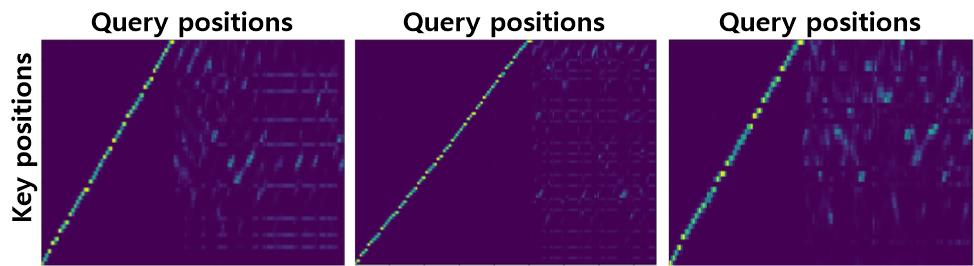}
		\label{attention2}
	}
\end{subfigure}
\caption{Visualization of the attention weights of the Oracle Teacher with and without the source input}
\label{attention}
\end{figure}

We trained another incomplete Oracle Teacher, called Oracle Teacher w/o source. 
The zero arrays, which had the same size of $x$, were treated as the input instead of the source input $x$ during the training. 
Then, the Oracle Teacher w/o source only considered the target input $y$ when making a prediction, similar to the aforementioned trivial solution.
In Fig. \ref{attention}, we visualize the cross attention scores
of the decoder for the ASR task, where the x-axis refers to acoustic frames and the y-axis refers to characters.
For the Oracle Teacher w/o source, the attention had almost diagonal alignment along with the key position (text) while ignoring the length of query, as shown in Fig. \ref{attention2}.
In contrast, the Oracle Teacher considered both speech and text alignment in the cross attention, and the attention scores were correctly computed along with the acoustic frames (query), as shown in Fig. \ref{attention1}.
It means that the Oracle Teacher utilized the source $x$ for model training while preventing the trivial solution.
Therefore, we can confirm that the Oracle Teacher, including SourceNet, has been correctly trained.  

\subsubsection{KD with Oracle Teacher w/o source}
In addition to the previous experiments, we proceeded to train the student model with the knowledge of the Oracle Teacher w/o source. However, as we expected, the distilled student failed to converge.
From this additional result, we can verify that the source input $x$ is the necessary factor of the Oracle Teacher, and the proposed Oracle Teacher has been correctly trained.

\subsubsection{Performance of Oracle Teacher} 
\label{per_oracle}
\begin{table}[t]
\centering
\fontsize{7.3}{8.76}\selectfont
\caption{Performance and parameter comparison between the Oracle Teacher and the conventional teacher.}
\begin{tabular}{|c|c|c|cccc|}
\hline
\multirow{3}{*}{Task} & \multirow{3}{*}{Model} & \multirow{3}{*}{Param.} & \multicolumn{4}{c|}{WER (\%)} \\ \cline{4-7} 
 &  &  & \multicolumn{2}{c|}{dev} & \multicolumn{2}{c|}{test} \\ \cline{4-7} 
 &  &  & \multicolumn{1}{c|}{clean} & \multicolumn{1}{c|}{other} & \multicolumn{1}{c|}{clean} & other \\ \hline
\multirow{2}{*}{ASR} & Jasper DR \cite{jasper:scheme}& 333 M & \multicolumn{1}{c|}{3.61} & \multicolumn{1}{c|}{11.37} & \multicolumn{1}{c|}{\textbf{3.77}} & 11.08 \\ \cline{2-7} 
 & Oracle Teacher & 33 M & \multicolumn{1}{c|}{\textbf{2.87}} & \multicolumn{1}{c|}{\textbf{3.10}} & \multicolumn{1}{c|}{4.03} & \textbf{3.29} \\ \hline
\end{tabular}
\label{oracle_perform}
\end{table}

We also evaluated the performance of the Oracle Teacher itself compared to the conventional teachers, as shown in Table \ref{oracle_perform}.
If the Oracle Teacher copies the target input $y$ without utilizing the source input $x$, the performance of the Oracle Teacher should be perfect.
We measured WER (\%) results on LibriSpeech.
The results show that the performance of the Oracle Teacher was more effective than that of the conventional teacher model, which seemed reasonable because the Oracle Teacher was trained with the guidance of $y$.
Meanwhile, the predictions of the Oracle Teacher were not the same as each ground truth. 
This implies that the Oracle teacher's output did not simply copy the target input $y$, and the information from a properly-trained SourceNet was utilized to generate the prediction.
Compared to the ``clean" datasets, the difference of WER was huge in ``other" sets.
Since the ``other" dataset represents a noisy dataset, the conventional ASR model (Jasper DR) showed low performance for the ``other" dataset. However, the Oracle Teacher could result in high performance for the noisy dataset since it used text information from the target.
This indicates that the Oracle Teacher could be a noisy robust teacher with small parameters.
From the results, we can verify that the Oracle Teacher provided more accurate and better guidance to the student than the conventional ASR teacher model.

\begin{table}[t]
\centering
\caption{performance comparison on LibriSpeech test datasets with greedy decoding. Conformer-T represents the Conformer-Transducer.}
{\fontsize{7.3}{8.76}\selectfont
\begin{tabular}{|c|c|cc|cc|}
\hline
\multirow{2}{*}{\begin{tabular}[c]{@{}c@{}}ASR baseline\\ model\end{tabular}} & \multirow{2}{*}{Params.} & \multicolumn{2}{c|}{WER (\%)} & \multicolumn{2}{c|}{RERR (\%)} \\ \cline{3-6} 
 &  & \multicolumn{1}{c|}{clean} & other & \multicolumn{1}{c|}{clean} & other\\ \hline
Conformer-T L & 120 M & \multicolumn{1}{c|}{2.31} & 5.02 & \multicolumn{1}{c|}{-} & - \\
Conformer-T S & 14 M & \multicolumn{1}{c|}{3.82} & 9.25 & \multicolumn{1}{c|}{-} & - \\ \hline \hline
\multirow{2}{*}{Student} & \multirow{2}{*}{Teacher} & \multicolumn{2}{c|}{WER (\%)} & \multicolumn{2}{c|}{RERR (\%)} \\ \cline{3-6} 
 & & \multicolumn{1}{c|}{clean} & other & \multicolumn{1}{c|}{clean} & other \\ \hline
\multirow{3}{*}{Conformer-T S} & None & \multicolumn{1}{c|}{3.82} & 9.25 & \multicolumn{1}{c|}{-} & -  \\
 & Conformer-T L & \multicolumn{1}{c|}{3.62} & 8.94 & \multicolumn{1}{c|}{5.24} & 3.35 \\
 & Oracle Teacher & \multicolumn{1}{c|}{3.51} & 8.65 & \multicolumn{1}{c|}{8.12} & 6.49 \\ \hline
\end{tabular}}
\label{conformer-transducer}
\end{table}
\begin{table}[t]
\centering
\caption{Computational resource consumption comparison across teacher models.}{%
{\fontsize{7.3}{8.76}\selectfont
\label{gpu_rnnt}
\begin{tabular}{|c|c|c|c|c|}
\hline
Teacher model & Params.& GPU & Batch & Epochs \\
\hline
Conformer-T L&120 M&128 * 50GB& 2048 & 200 epochs \\
\textbf{Oracle Teacher (ours)}&16 M&4 * 45GB& 64 & 25 epochs\\
\hline
\end{tabular}%
}}
\end{table}

\subsection{Application to Transducer Framework}
In previous experiments, we mainly focused on the distillation for CTC models. To check the applicability of the proposed method to the RNN-T model, we conducted a new distillation scenario, where the teacher and student were based on the RNN-T framework.
Since the RNN-T also adopted the many-to-one alignment process, we could easily apply the proposed method to the Transducer model.
As for the conventional teacher, we adopted a large Conformer-Transducer (Conformer-T L) model, the current SOTA for ASR. 
The RNN-T student was based on a small Conformer model (Conformer-T S), where the encoder network consisted of 16 layers with 176 dimensions.
In the proposed framework, the encoder network of the RNN-T model was replaced with the Oracle Teacher.
By using both acoustic and linguistic information, the Oracle Teacher learned the unified representation.
Then, the joint network integrated the outputs of the Oracle Teacher and prediction network in generating the prediction.
In this experiment, the SourceNet had the same architecture as the encoder network of the RNN-T student model, consisting of 16 Conformer layers.
When distilling the knowledge, we used FitNets as the KD technique.
As shown in Table \ref{conformer-transducer}, it is verified that the student distilled from the proposed teacher model was better than the student distilled from the conventional Transducer teacher (Conformer-T L).
While the student distilled from Conformer-T L produced WER 3.62 \%/8.94 \% on test-clean and test-other, the student distilled from the Oracle Teacher yielded 3.51 \%/8.65 \% on test datasets.
This means that the Oracle Teacher was more supportive in distilling the knowledge to the RNN-T student.
Also, as reported in Table \ref{gpu_rnnt}, our framework dramatically reduced the computational cost to train the teacher model.
The training of Conformer-T L required 128 GPUs for 200 epochs with a batch size of 2048.
Compared to the conventional RNN-T teacher model, the proposed method could achieve meaningful results with relatively limited resources in the RNN-T framework.

\begin{table}[t]
\centering
\caption{Performance comparison on LibriSpeech test datasets with greedy decoding.}
{\fontsize{7.3}{8.76}\selectfont
\begin{tabular}{|ccc|cc|}
\hline
\multicolumn{3}{|c|}{\multirow{2}{*}{\begin{tabular}[c]{@{}c@{}}ASR baseline\\ model\end{tabular}}}                                                                                                                                                                  & \multicolumn{2}{c|}{WER}           \\ \cline{4-5} 
\multicolumn{3}{|c|}{}                                                                                                                                                                                                                                               & \multicolumn{1}{c|}{clean} & other \\ \hline
\multicolumn{3}{|c|}{Jasper DR}                                                                                                                                                                                                                                      & \multicolumn{1}{c|}{3.77}  & 11.08 \\ \hline
\multicolumn{3}{|c|}{Jasper Mini}                                                                                                                                                                                                                                    & \multicolumn{1}{c|}{8.85}  & 24.26 \\ \hline
\multicolumn{1}{|c|}{\multirow{2}{*}{Student}}                                               & \multicolumn{1}{c|}{\multirow{2}{*}{Teacher}}                                                  & \multirow{2}{*}{Target input}                                        & \multicolumn{2}{c|}{WER}           \\ \cline{4-5} 
\multicolumn{1}{|c|}{}                                                                       & \multicolumn{1}{c|}{}                                                                          &                                                                      & \multicolumn{1}{c|}{clean} & other \\ \hline
\multicolumn{1}{|c|}{\multirow{3}{*}{\begin{tabular}[c]{@{}c@{}}Jasper\\ Mini\end{tabular}}} & \multicolumn{1}{c|}{\multirow{3}{*}{\begin{tabular}[c]{@{}c@{}}Oracle\\ Teacher\end{tabular}}} & Ground truth                                                         & \multicolumn{1}{c|}{6.67}  & 19.82 \\ \cline{3-5} 
\multicolumn{1}{|c|}{}                                                                       & \multicolumn{1}{c|}{}                                                                          & \begin{tabular}[c]{@{}c@{}}Pseudo label (Jasper DR)\end{tabular}   & \multicolumn{1}{c|}{6.68}  & 19.85 \\ \cline{3-5} 
\multicolumn{1}{|c|}{}                                                                       & \multicolumn{1}{c|}{}                                                                          & \begin{tabular}[c]{@{}c@{}}Pseudo label (Jasper Mini)\end{tabular} & \multicolumn{1}{c|}{6.89}  & 20.05 \\ \hline
\end{tabular}}
\label{ps_result}
\end{table}

\subsection{Training Oracle Teacher with Pseudo Label} 
This work offered a powerful but efficient teacher model in the KD framework. 
However, using ground truth transcription in KD could limit its usage on unsupervised data. 
One possible way to apply the Oracle Teacher in the unsupervised setting is leveraging the pseudo label instead of the ground truth.
To check the feasibility, extending the proposed framework further, we conducted a new KD scenario, where the pseudo label was used in the proposed framework instead of the ground truth. 
We extracted the pseudo label from the two Jasper DR and Jasper Mini models. 
In the previous experiment, the Jasper DR and Jasper Mini were adopted as the conventional ASR teacher and the student baseline, respectively. 
We utilized the pseudo label instead of the ground truth during the teacher training and knowledge extraction stages. 
Note that the pseudo label was employed as both additional input and target of the Oracle Teacher during the teacher training.
Interestingly, as reported in Table \ref{ps_result}, we confirmed that using the pseudo label was possible with the proposed method, being robust to the error in the ground truth.
When using the pseudo label from the Jasper DR, there was no significant performance difference from the original Oracle Teacher since the pseudo label might be similar to the ground-truth.
In the case of leveraging the prediction from the Jasper Mini, the corresponding Oracle Teacher was worse than the others, implying that the error in the pseudo label affected the WER performance of the student model.
However, it still worked well in distilling the knowledge, providing WER 6.89 \% and 20.05 \% on test-clean and test-other, respectively.
From the results, it is confirmed that we could use the Oracle Teacher with the unlabeled dataset, like self-training.

\section{Conclusions}
We introduced a novel teacher for CTC-based sequence models, namely Oracle Teacher, that leverages the output labels as the additional input to the model.
Through a number of experiments, we confirmed that the student distilled from the Oracle Teacher performed better compared to the one distilled from the conventional teacher.
Furthermore, our framework significantly reduced the computational cost of the teacher model in terms of the training time and required GPU resources.
As the effective teacher can be trained with a reduced computational cost, the Oracle Teacher can be a new breakthrough in KD.
We also explored the feasibility of applying the Oracle Teacher approach to different applications, including the transducer framework and the self-training-like setting that utilized pseudo labels instead of ground-truth labels. 
We expect the application of the Oracle Teacher in various tasks, such as regression, ranking, etc., in the future.


\ifCLASSOPTIONcaptionsoff
  \newpage
\fi

\bibliographystyle{unsrt}
\bibliography{Trans.bib}

\end{document}